\pdfoutput=1

\documentclass[11pt]{article}

\usepackage[preprint]{acl}

\usepackage{times}
\usepackage{amsmath}
\usepackage{amsfonts}
\usepackage{bm}
\usepackage{mathrsfs}
\usepackage{bbm}
\usepackage{nicefrac}
\usepackage{empheq}
\usepackage{diagbox}

\usepackage{graphicx}
\usepackage{booktabs}
\usepackage{multirow}
\usepackage{multicol}
\usepackage{array}
\usepackage{makecell}
\usepackage{bigstrut}
\usepackage{diagbox}
\usepackage{threeparttable}
\usepackage{xcolor,colortbl}
\usepackage{tabularx}

\usepackage{algorithm}
\usepackage{algpseudocode}

\usepackage{xspace}
\usepackage{mfirstuc}
\usepackage{tabulary}
\usepackage{framed}
\usepackage[normalem]{ulem}
\useunder{\uline}{\ul}{}

\usepackage{enumitem}

\usepackage{tikz}
\usepackage{tcolorbox}
\usepackage{pifont}

\definecolor{mainboxbg}{HTML}{F7F9FC}
\definecolor{mainboxborder}{HTML}{A1C6EA}
\definecolor{BoxBackground}{RGB}{240, 240, 240} % 浅灰色背景
\definecolor{BoxFrame}{RGB}{0, 0, 0} % 黑色边框
\definecolor{TitleBackground}{RGB}{0, 0, 0} % 标题背景颜色
\definecolor{TitleText}{RGB}{255, 255, 255} % 标题文字颜色
\definecolor{lightyellow}{rgb}{1.0, 1.0, 0.6} 
\definecolor{lightcyan}{rgb}{0.48, 1.0, 1.0}

\newtcbox{\mybox}[1][red]
  {on line, arc = 0pt, outer arc = 0pt,
    colback = #1!10!white, colframe = #1!50!black,
    boxsep = 0pt, left = 1pt, right = 1pt, top = 2pt, bottom = 2pt,
    boxrule = 0pt, bottomrule = 1pt, toprule = 1pt}

\tcbset{
  academicbox/.style={
    boxsep=5pt,
    left=2pt,
    right=2pt,
    bottom=0.5pt,
    boxrule=0.5pt,
    colback=BoxBackground,
    colframe=BoxFrame,
    colbacktitle=TitleBackground,
    coltitle=TitleText,
    enhanced,
    attach boxed title to top left={yshift=-0.1in,xshift=0.1in},
    boxed title style={boxrule=0pt,colframe=white},
    title={#1},
  }
}

\newtcolorbox{AcademicBox}[1][]{academicbox=#1}

\colorlet{mythmback}{gray!40!white}
\newtcolorbox{boxEnv}{
colback=mythmback,coltitle=gray,colframe=mythmback,
center,
width=\linewidth,
boxrule=0.1pt,
left=1pt,right=1pt,
top=2pt,bottom=2pt,
before skip=10pt, after skip=10pt,
fontupper=\footnotesize, % Adjust the font size here
}

\newtcolorbox{caseBoxEnv}{
colback=white,colframe=black,
center,
width=\linewidth,
boxrule=0.1pt,
left=1pt,right=1pt,
top=2pt,bottom=2pt,
}

\def\method{{\sc EvolSQL}\xspace}
\def\textbfmethod{{\sc \textbf{EvolSQL}}\xspace}

\title{\textbfmethod: Structure-Aware Evolution for Scalable \\ Text-to-SQL Data Synthesis}

\author{Xuanguang Pan$^{\clubsuit}$, \textbf{Chongyang Tao}$^{\clubsuit}$,\ \   Jiayuan Bai$^{\clubsuit}$, Jianling Gao$^{\clubsuit}$, \textbf{Zhengwei Tao}$^{\spadesuit}$, \\  \textbf{Xiansheng Zhou}$^{\bigtriangleup}$, \ \ \textbf{Gavin Cheung}$^{\bigtriangleup}$, \ \ \textbf{Ma Shuai}$^{\clubsuit}$ \\
 $^{\clubsuit}$ SKLCCSE Lab, Beihang University  \quad 
$^{\spadesuit}$Peking University  
$^{\bigtriangleup}$Independent Researcher
\\
\texttt{\{panxg,chongyang,baijiayuan,jianlingg,mashuai\}@buaa.edu.cn} \\
 \\
}

\begin{document}
\maketitle
\begin{abstract}

Training effective Text-to-SQL models remains challenging due to the scarcity of high-quality, diverse, and structurally complex datasets. Existing methods either rely on limited human-annotated corpora, or synthesize datasets directly by simply prompting LLMs without explicit control over SQL structures, often resulting in limited structural diversity and complexity. To address this, we introduce {\sc EvolSQL}, a structure-aware data synthesis framework that evolves SQL queries from seed data into richer and more semantically diverse forms. {\sc EvolSQL} starts with an \emph{exploratory Query-SQL expansion} to broaden question diversity and improve schema coverage, and then applies an \emph{adaptive directional evolution} strategy using six \emph{atomic transformation operators} derived from the SQL Abstract Syntax Tree to progressively increase query complexity across relational, predicate, aggregation, and nesting dimensions. An execution-grounded SQL refinement module and schema-aware deduplication further ensure the creation of high-quality, structurally diverse mapping pairs. Experimental results show that a 7B model fine-tuned on our data outperforms one trained on the much larger SynSQL dataset using only 1/18 of the data.

\end{abstract}

\section{Introduction}
\label{Introduction}

The task of Text-to-SQL aims to translate natural language questions into executable SQL queries, enabling non-expert users to interact with complex databases using everyday language~\citep{fu2023catsql}. As a core interface between human intent and structured data systems, it has become increasingly important in real-world applications such as business analytics, scientific data exploration, and enterprise search.
Recent advances in large language models (LLMs) have substantially improved Text-to-SQL performance, positioning LLMs as the dominant backbone for modern systems~\citep{li2024dawn}. Despite this progress, robust generalization to unseen schemas and complex query structures remains a central challenge.

Current approaches to this task fall into two paradigms. Multi-agent frameworks~\citep{pourreza2023din, talaei2024chess, wang2025mac}. These methods improve reasoning and schema grounding without additional model training, and can yield noticeable gains on challenging benchmarks. 
However, approaches built on closed-source models suffer from inherent drawbacks such as data privacy, cost, and deployment flexibility.

Thus, recent research has increasingly shifted toward training-based paradigms built on open-source models, including supervised fine-tuning and reinforcement learning, aiming to specialize models for robust Text-to-SQL generation~\citep{li2024codes, pourreza2025reasoning}.

The advancement of training-based approaches is fundamentally constrained by the availability and quality of training data. While human-annotated datasets such as Spider~\citep{yu2018spider} and BIRD~\citep{li2024can} provide high-fidelity  pairs, their scale and structural diversity remain limited. While direct prompting~\citep{yang2024synthesizing} increases scale, it often struggles with structural diversity and logical consistency.  Alternatively, recent works like OmniSQL~\citep{li2025omnisql} synthesize massive datasets from web tables, but achieving competitive performance requires millions of samples, leading to high computational costs.

We propose \method, a structure-aware data synthesis framework that systematically evolves SQL queries from simple seeds into structurally richer and semantically diverse forms. \method begins with an \emph{exploratory Query-SQL expansion} stage, which broadens query intents and enhances schema coverage by explicitly referencing under-explored elements. 
Building upon this foundation, we define six \emph{atomic transformation operators}
that manipulate distinct structural dimensions, namely functional wrapping, operator mutation, logical clause expansion, relational expansion, nesting evolution, and set composition. These operators are orchestrated via an \emph{adaptive directional evolution} strategy, where transformations are guided by the current query structure and schema context rather than applied randomly.
Additionally, we incorporate an execution-grounded SQL refinement and a schema-aware deduplication module in the synthesis process. These components ensure the generation of high-quality pairs while maintaining structural diversity, further enhancing the utility of the synthesized dataset for downstream model training.

\begin{table}[t!]
  \centering
  \scriptsize
  \setlength{\tabcolsep}{2pt}
  \resizebox{0.48\textwidth}{!}{%
  \begin{tabular}{lccccc}
    \toprule
    \textbf{Method} & 
    \begin{tabular}[c]{@{}c@{}}\textbf{Synthetic}\\ \textbf{Schema}\end{tabular} & 
    \begin{tabular}[c]{@{}c@{}}\textbf{Atom.}\\ \textbf{Control}\end{tabular} & 
    \begin{tabular}[c]{@{}c@{}}\textbf{Prog.}\\ \textbf{Cmplx.}\end{tabular} & 
    \begin{tabular}[c]{@{}c@{}}\textbf{Refine.}\\ \textbf{Mechanism}\end{tabular} & 
    \begin{tabular}[c]{@{}c@{}}\textbf{CoT}\\ \textbf{Trace}\end{tabular} \\
    \midrule
    SENSE~\citep{yang2024synthesizing} & \ding{51} & \ding{55} & \ding{55} & \ding{55} & \ding{55} \\
    OmniSQL~\citep{li2025omnisql} & \ding{51} & \ding{55} & \ding{55} & \ding{55} & \ding{51} \\
    SQLFLOW~\citep{cai2025text2sql} & \ding{55} & \ding{55} & \ding{55} & \ding{55} & \ding{51} \\
    \midrule
    \textbf{\method} & \ding{55} & \ding{51} & \ding{51} & \ding{51} & \ding{51} \\ 
    \bottomrule
  \end{tabular} 
  }
  \vspace{-2mm}
  \caption{Comparison of synthesis frameworks. Atom. Control.: Atomic-level Control; Prog. Cmplx.: Progressive Complexity.}
  \label{tab:method_comparison}
  \vspace{-4mm}
\end{table}

To evaluate the effectiveness of \method, we fine-tune a 7B model on the synthesized dataset. On the BIRD development set, the model achieves an execution accuracy of 65.1\%, outperforming a model of the same scale trained on the much larger SynSQL dataset~\citep{li2025omnisql}, despite using only approximately 1/18 of the training data. Furthermore, the model demonstrates strong generalization capabilities on benchmarks not involved in our augmentation process.  To summarize, our contributions are fourfold:
\begin{itemize}[leftmargin=8pt, itemsep=0pt, topsep=2pt, partopsep=0pt]
    \item We propose \method, a fully automated framework for Text-to-SQL dataset synthesis that explicitly models and controls SQL structural properties. 
    
    \item We design a family of \emph{atomic transformation operators} and an \emph{adaptive directional evolution} strategy. By decomposing SQL complexity into atomic mutations, this mechanism systematically scales query complexity from exploratory seeds while maintaining high logical rigor.
    
    \item We introduce an execution-grounded refinement module and schema-aware deduplication, which collectively promote the quality and semantic diversity of generated data.
    
    \item  Experiments show \method significantly boosts Text-to-SQL performance, surpassing recent data synthesis baselines on BIRD with only 1/18 of the training samples.

\end{itemize}

\section{Related Works} \label{related works}

\vspace{-1mm}
\paragraph{Text-to-SQL Generation.}
% Early Research
Early Text-to-SQL approaches primarily relied on rule-based methods or neural sequence-to-sequence architectures~\citep{basik2018dbpal,sun2018semantic,wang2020rat}. However, these methods often struggled with complex queries and demonstrated limited cross-domain generalization. The emergence of LLMs has fundamentally transformed the field, providing substantially improved reasoning and generalization capabilities~\citep{pourreza2023din, gao2024text,liu2023comprehensive,dong2023c3}. Building on these capabilities, recent methods have moved beyond single-pass generation, adopting multi-agent frameworks that decompose the task into specialized sub-stages, such as schema linking, self-correction, and candidate selection~\citep{pourreza2024chase,talaei2024chess,wang2025mac,gao2024xiyansql}.

% training
Beyond multi-agent pipelines, training strategies have also evolved to specialize models for the Text-to-SQL domain. Supervised fine-tuning (SFT) for domain-specific instruction alignment enables open-source models to achieve competitive performance~\citep{pourreza2024dts, he2025starSQL}. 
To push performance boundaries, reinforcement learning techniques have been employed to further enhance the model's reasoning capabilities~\citep{liu2025dpo, zhai2025excot, pourreza2025reasoning, yao2025arctic}.

\vspace{-1mm}
\paragraph{Text-to-SQL Data Synthesis.}
% early work
Developing robust Text-to-SQL models is often hindered by the narrow coverage of available datasets, leading to the exploration of diverse data synthesis strategies.
Traditional synthesis often relied on probabilistic grammars or templates to generate pairs \citep{wang2021learning, wu2021data, guo2018question}, or converted synthetic questions into SQL \citep{yang2021hierarchical, weir2020dbpal}. However, these methods were either constrained by rigid templates or suffered from semantic noise and logical mismatches.

Recent studies leverage the generative capabilities of LLMs to scale data synthesis beyond template-based constraints. While \citet{yang2024synthesizing} directly generate data using LLMs, such single-pass prompting lacks rigorous verification.
To expand schema variety, \citet{li2025omnisql} construct the massive SynSQL-2.5M dataset from web sources. While providing extensive domain coverage, this approach suffers from low data efficiency, requiring millions of samples to achieve significant gains.
Most recently, \citet{cai2025text2sql} propose a framework employing diverse augmentation strategies.
However, as complexity enhancement is treated as merely one of several dimensions, the process lacks a systematic direction, hindering the progressive scaling of query difficulty. In contrast, \method leverages an adaptive directional evolution strategy to provide a structured and scalable path for progressively elevating data complexity. Table~\ref{tab:method_comparison} provides a detailed comparison between \method and these existing synthesis frameworks.
\section{Problem Formalization}
\label{sec:formalization}

\paragraph{Text-to-SQL.} We define a Text-to-SQL instance as a triplet $(q, s, \mathcal{S})$, where $q$ represents a natural language query, $s$ denotes the corresponding SQL logic, and $\mathcal{S}$ represents the database schema,  providing the structural context including table definitions, column attributes, and relational schema constraints. The task aims to learn a mapping $f: (q, \mathcal{S}) \to s$ that accurately translates user intents into executable SQL queries.

\paragraph{Text-to-SQL Data Synthesis.} This task aims to automatically construct high-quality instances $\mathcal{D}_{syn} = \{(q', s', \mathcal{S})\}$ either from scratch or by expanding an initial dataset $\mathcal{D}_{seed}$. 
This synthesis process seeks to increase both the diversity of query intents and the coverage of schema elements, while ensuring that the generated SQL remains executable and grounded in the database.

\section{Method}
\label{sec:method}

\begin{figure*}[t!]
    \centering
    % \vspace{}
    \vspace{-4mm}
    \includegraphics[width=0.99\linewidth]{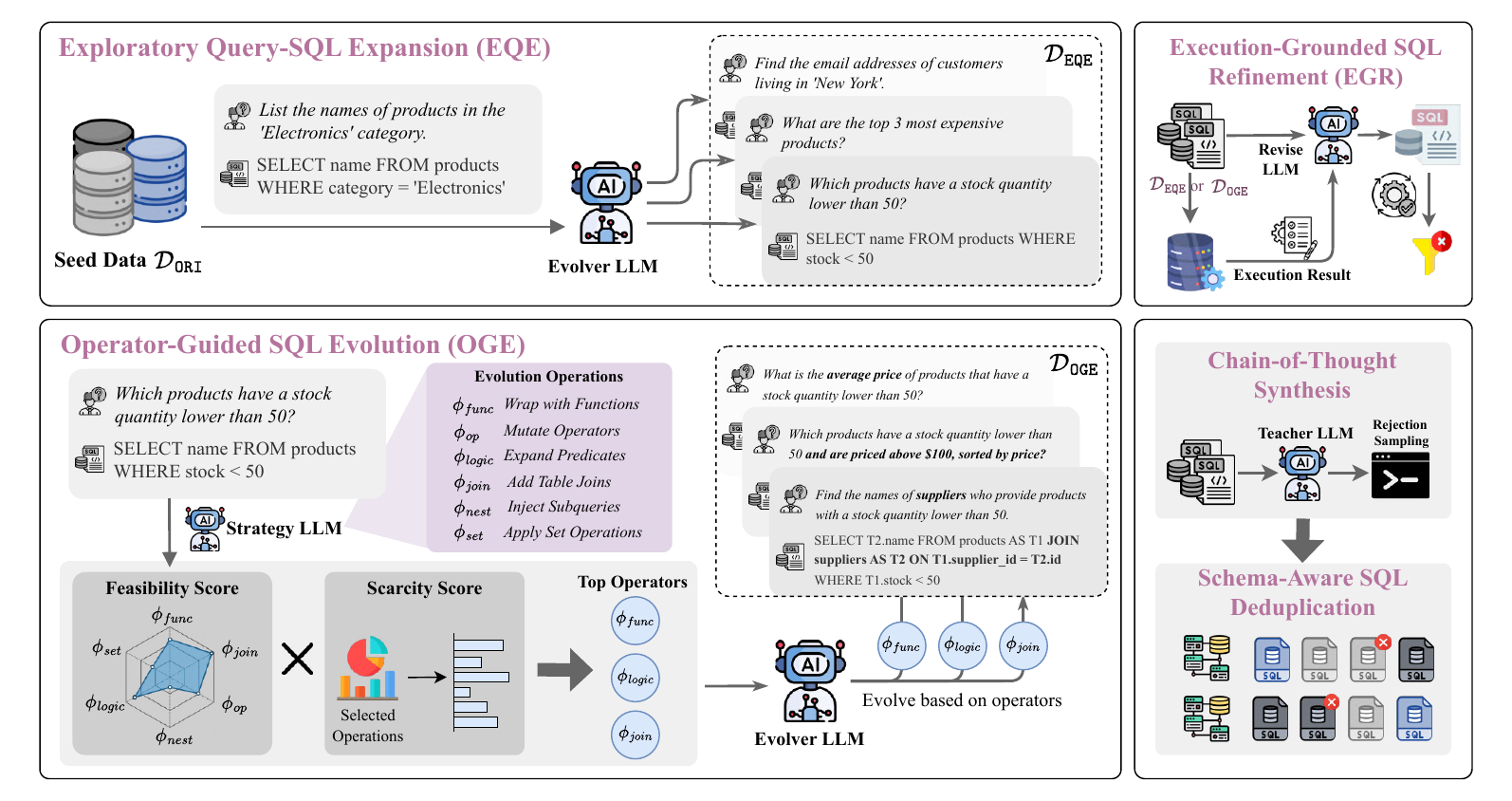}
    \vspace{-3mm}
    \caption{Overview of the \method data synthesis pipeline.}
    \vspace{-4.5mm}
    \label{fig:pipeline}
\end{figure*}

As illustrated in Figure~\ref{fig:pipeline}, our framework synthesizes the dataset through a progressive evolution pipeline. We begin with \emph{Exploratory Query-SQL Expansion (EQE)} (Sec.~\ref{sec:horizontal}), which expands the semantic scope of user intents and enhances database schema coverage. Building on this foundation, we proceed to \emph{Operator-Guided SQL Evolution (OGE)} (Sec.~\ref{sec:vertical}), where we utilize a family of \emph{atomic transformation operators} to systematically increase structural complexity along orthogonal dimensions. Finally, we perform \emph{Chain-of-Thought Solution Synthesis} (Sec.~\ref{sec:augcot}), ensuring high data quality and diversity through execution-verified Chain-of-Thought (CoT) synthesis and schema-aware deduplication.

\subsection{Exploratory Query-SQL Expansion}
\label{sec:horizontal}

Existing Text-to-SQL benchmarks often suffer from sparse schema utilization due to finite sample sizes and the difficulty of manual annotation.
This leaves significant portions of tables and relational dependencies under-explored. To bridge this gap, we propose to systematically synthesize diverse queries that leverage these under-utilized components.
In practice, we prompt an LLM $\mathcal{M}_{gen}$ to draw inspiration from a given NL2SQL example $(q,s,\mathcal{S})$ and jointly generate a novel, semantically coherent natural language query $\tilde{q}$ and its corresponding SQL draft $\tilde{s}$, such that the generated query plausibly maps to a valid SQL statement. Formally, this process is expressed as:
\vspace{-1mm}
    \begin{equation}
    \vspace{-1mm}
        (\tilde{q}, \tilde{s}) \leftarrow \mathcal{M}_\texttt{gen}(q, s,\mathcal{S}; \mathcal{I})
    \end{equation}
where $\mathcal{I}$ is an evolution instruction. This mechanism encourages both novelty in query intent and coverage of diverse schema elements. However, the resulting SQL drafts may still contain syntax errors, logical inconsistencies, or incomplete schema grounding. We then apply an \emph{execution-grounded SQL refinement} module, using execution feedback to refine the SQL. Specifically, we execute the candidate SQL $\tilde{s}$ against the database $\mathcal{DB}$ to obtain feedback $r = \texttt{Exec}(\tilde{s}, \mathcal{DB})$. Whether $r$ is an error message or an execution result, we feed it into a correction module to refine the SQL, ensuring executability and data grounding:
\vspace{-1mm}
    \begin{equation}
    \vspace{-1mm}
        s' \leftarrow \mathcal{M}_\texttt{refine}(\tilde{q}, \tilde{s}, \mathcal{S}; r)
    \end{equation}
The final output is retained as $(q', s') = (\tilde{q}, s')$ only if $s'$ executes successfully and yields a non-empty result. Through this process, we construct an expanded dataset $\mathcal{D}_{H}$, providing a diverse and grounded foundation that offers a rich variety of starting points for subsequent evolution in query complexity and depth. 

\subsection{Operator-Guided SQL Evolution}
\label{sec:vertical}

Building upon the diverse data collected in the Exploratory Query-SQL Expansion stage, we introduce \emph{operator-guided SQL evolution} to systematically increase the reasoning complexity of generated queries. The goal of this stage is to construct substantially harder SQL queries by progressively enriching their logical structure, including more intricate conditions, nested subqueries, and compositional interactions among operators.

Rather than simply applying random modifications, this process requires structured and controllable evolution of query logic. We formalize SQL complexity through the topology of its Abstract Syntax Tree (AST), which provides an explicit representation of SQL operators and their hierarchical relationships. Let $\mathcal{T}$ denote an AST,  
and represent any subtree rooted at node $v$ as a tuple
$\mathcal{T}_v = \langle \ell, \mathcal{C} \rangle$, where $\ell$ is the node label (e.g., \texttt{SELECT}, \texttt{AND}) and $\mathcal{C}$ the ordered set of children. We then define a family of transformation operators $\Phi$ to evolve AST along different dimensions:

\vspace{1mm}
\noindent \textbf{\ding{182} Functional Wrapping ($\phi_\texttt{func}$):}
    Wraps a leaf node in a function to increase local expression complexity. Given a leaf $\mathcal{T}_\texttt{leaf} = \langle \ell_\texttt{col}, \emptyset \rangle$, it transforms to:
    \vspace{-1mm}
    \begin{equation}
    \vspace{-1mm}
        \phi_\texttt{func}(\mathcal{T}_\texttt{leaf}) = \langle f, \{ \mathcal{T}_\texttt{leaf} \} \rangle
        \vspace{-1mm}
    \end{equation}
    \vspace{-1mm}
    where $f$ is a function label (e.g., \texttt{AVG}, \texttt{YEAR}).

\vspace{1mm}
\noindent \textbf{\ding{183} Operator Mutation ($\phi_\texttt{op}$):}
    Embeds a simple expression into a complex operator. Let $\Omega$ be a set of complex operators (e.g., \texttt{CASE}, \texttt{BETWEEN}). For an expression subtree $\mathcal{T}_{expr}$, it constructs:
    \vspace{-1mm}
    \begin{equation}
    \vspace{-1mm}
    \phi_\texttt{op}(\mathcal{T}_\texttt{expr}) = \langle \omega, \mathcal{C}_\texttt{new} \rangle
    \vspace{-1mm}
    \end{equation}
    where $\mathcal{T}_\texttt{expr} \in \mathcal{C}_\texttt{new}$ and $\omega$ is a new operator label.

\vspace{1mm}
\noindent \textbf{\ding{184} Logical Clause Expansion ($\phi_\texttt{logic}$):} 
    Enhances the width of clause nodes $v \in $ $\{ \texttt{WHERE}, $ $ \texttt{HAVING}, \texttt{ORDER BY} \}$by adding constraint $e_\texttt{new}$ via a connector $\lambda$:
    \vspace{-1mm}
    \begin{equation}
     \vspace{-1mm}
        \phi_\texttt{logic}(v) = \langle v, \text{Comb}_{\lambda}(\mathcal{C}_v, e_\texttt{new}) \rangle
         \vspace{-1mm}
    \end{equation}
where $\mathcal{C}_v$ denotes the set of children of $v$, and $\text{Comb}_{\lambda}$ is the combination function.

\vspace{1mm}
\noindent \textbf{\ding{185} Relational Expansion ($\phi_\texttt{join}$): }
    Increases relational complexity. Given $\mathcal{T}_\texttt{from} = \langle \text{FROM}, \mathcal{C} \rangle$, it appends a join subtree $\mathcal{T}_{sub} = \langle \text{JOIN}, \{ T_\texttt{new}, \text{cond} \} \rangle$:
    \vspace{-1mm}
    \begin{equation}
    \vspace{-1mm}
        \phi_\texttt{join}(\mathcal{T}_\texttt{from}) = \langle \text{FROM}, \mathcal{C} \cup \{ \mathcal{T}_{sub} \} \rangle
     \vspace{-1mm}
    \end{equation}
    $T_{new}$ is the new table and $\text{cond}$ the join condition.

\vspace{1mm}
\noindent \textbf{\ding{186} Nesting Evolution ($\phi_\texttt{nest}$):}
    Increases tree depth by replacing a leaf node with a recursive query structure. For a value node $\mathcal{T}_{val} = \langle \text{value}, \emptyset \rangle$, it performs the substitution:
    \vspace{-1mm}
    \begin{equation}
        \phi_\texttt{nest}(\mathcal{T}_\texttt{val}) = \mathcal{T}_\texttt{sub}
        \vspace{-1mm}
    \end{equation}
    where $\mathcal{T}_\texttt{sub}$ represents a complete, independent subquery tree.

\vspace{1mm}
\noindent \textbf{\ding{187} Set Composition ($\phi_\texttt{set}$):}
    Combine two independent query trees to form a compound structure. Given the tree $\mathcal{T}$, it constructs a new root node:
    \vspace{-1mm}
    \begin{equation}
        \phi_{set}(\mathcal{T}) = \langle \odot, \{ \mathcal{T}, \mathcal{T}_\texttt{new} \} \rangle
        \vspace{-1mm}
    \end{equation}
    where $\odot \in \{ \text{UNION}, \text{INTERSECT}, \text{EXCEPT} \}$ denotes the set operator.

While $\Phi$ defines the possible directions of evolution, naïvely selecting operators at random is insufficient for constructing a high-quality dataset. Such an approach suffers from two fundamental issues: (i) \emph{structural invalidity}, where certain transformations are incompatible with the current AST state (e.g., applying $\phi_{\texttt{nest}}$ to a query without eligible leaf nodes), and (ii) \emph{distributional bias}, where simpler operators (e.g., $\phi_{\texttt{logic}}$) are repeatedly favored, leading to mode collapse and limited structural diversity.

To address these challenges, we propose an \emph{adaptive directional strategy} that selects evolution directions based on both local feasibility and global diversity. Unlike passive filtering approaches that prune invalid samples post-generation, our strategy acts as an efficient pre-judgment mechanism to guide the evolution process. Specifically, for a given instance $(q, s)$, we evaluate each candidate operator $\phi \in \Phi$ using two complementary metrics: a \emph{feasibility score} and a \emph{scarcity weight}. First, to assess whether a transformation is structurally applicable, we define a \emph{feasibility score} $S_{\texttt{feas}}(\phi)$. A strategy model $\mathcal{M}_{\texttt{strat}}$ analyzes the current query and predicts the applicability of the atomic mutations, approximating structural constraints without explicit rules:
\vspace{-1mm}
\begin{equation}
\vspace{-1mm}
    S_{\texttt{feas}}(\phi) \leftarrow \mathcal{M}_{\texttt{strat}}(q, s, S; \phi)
    \vspace{-1mm}
\end{equation}

Second, to counteract operator imbalance, we introduce a \emph{scarcity weight} $W_{\texttt{div}}(\phi)$, which dynamically prioritizes under-represented evolution directions. Let $P_{\texttt{accum}}(\phi)$ denote the proportion of operator $\phi$ accumulated so far, and $P_{\texttt{target}}(\phi)$ the desired distribution (e.g., uniform). The scarcity weight is defined as:
\vspace{-1mm}
\begin{equation}
\vspace{-1mm}
    W_{\texttt{div}}(\phi) = \frac{P_{\texttt{target}}(\phi)}{P_{\texttt{accum}}(\phi) + \epsilon},
    \vspace{-1mm}
\end{equation}
where $\epsilon$ is a smoothing constant. This formulation encourages exploration of less frequent operators, thereby maintaining balanced structural coverage. Finally, we integrate the two metrics to compute a joint utility score, defined as:
\begin{equation}
    U(\phi) = S_{\texttt{feas}}(\phi) \cdot W_{\texttt{div}}(\phi),
\end{equation}
which adaptively shifts the evolution focus as the dataset grows. We select the top-$K$ operators $\{\phi^*_1, \dots, \phi^*_K\}$ with the highest utility and apply the corresponding evolution via the expansion and refinement module described in Sec.~\ref{sec:horizontal}:
\vspace{-1mm}
\begin{equation}
    \begin{aligned}
        (\tilde{q}, \tilde{s}) &\leftarrow \mathcal{M}_\texttt{gen}(q, s, \mathcal{S}; \mathcal{I}_{\phi^*}) \\
        s' &\leftarrow \mathcal{M}_\texttt{refine}(\tilde{q}, \tilde{s}, \mathcal{S}; r)
    \end{aligned}
    \vspace{-1mm}
\end{equation}
Each newly generated instance $(q', s')$ then serves as the seed for subsequent evolution rounds. This iterative process progressively expands the dataset toward higher-complexity. The full process is summarized in Algorithm~\ref{alg:vertical_evo} in Appendix~\ref{app:alg}.

\subsection{Chain-of-Thought Solution Synthesis}
\label{sec:augcot}

Chain-of-Thought (CoT) reasoning has proven instrumental in tackling complex tasks by decomposing them into intermediate logical steps. 
To harness this capability, we leverage a teacher LLM to synthesize CoT solutions via rejection sampling. 
For each instance $(q, s, \mathcal{S}) \in \mathcal{D} \cup \mathcal{D}_{syn}$, we sample $n$ independent candidate pairs $\{(c^{(i)}, \hat{s}^{(i)})\}_{i=1}^n$, each consisting of a reasoning trace and a predicted SQL. 
To ensure reliability, we validate these candidates by executing each $\hat{s}^{(i)}$ and comparing the result with that of the gold SQL $s$. If at least one candidate is correct, we retain the instance and attach the successful reasoning trace $c^{(t^*)}$ to it; otherwise, the instance is discarded. This execution-verified process yields the final training set $\mathcal{D}_{cot}$.

\begin{table*}[t!]
\setlength{\abovecaptionskip}{0.0cm}
\vspace{-2mm}
\begin{center}
\resizebox{0.9\textwidth}{!}{%
\begin{tabular}{l c ccc cc}\toprule
\multirow{2}{*}{Methods}&\multirow{2}{*}{\# Samples}&\multicolumn{2}{c}{BIRD}&\multicolumn{3}{c}{Spider}\\\cmidrule(lr){3-4} \cmidrule(lr){5-7}
& &Dev-EX&Dev-VES&Dev-EX&Dev-TS&Test-EX\\\midrule
\multicolumn{7}{l}{\emph{Prompting with Proprietary LLMs}}\\
GPT-4~\citep{achiam2023gpt} & - & 46.4 & 49.8 & 72.9 & 64.9 & - \\
DIN-SQL + GPT-4~\citep{pourreza2023din} & - & 50.7 & 58.8 & 82.8 & 74.2 & 85.3 \\
DAIL-SQL + GPT-4~\citep{gao2024text} & - & 54.8 & 56.1 & 83.5 & 76.2 & 86.6 \\
MAC-SQL + GPT-4~\citep{wang2025mac} & - & 59.4 & 66.2 & 86.8 & - & 82.8 \\
MCS-SQL + GPT-4~\citep{lee2024mcs} & - & 63.4 & 64.8 & 89.5 & - & 89.6 \\

\midrule
\multicolumn{7}{l}{\emph{Prompting with Open-Source LLMs}}                 \\
Llama3-8B~\citep{touvron2023llama} & - & 32.1 & 31.6 & 69.3 & 58.4 & 69.1 \\
Llama-3.1-8B-Instruct~\citep{grattafiori2024llama} & - & 42.0 & 40.8 & 71.9 & 61.8 & 72.2\\ 
Qwen2.5-7B~\citep{yang2024qwen2} & - & 41.1 & 42.0 & 72.5 & 64.0 & 75.9 \\
Qwen2.5-Coder-7B-Instruct~\citep{yang2024qwen2} & - & 50.9 & 48.3 & 79.1 & 73.4 & 82.2 \\

DIN-SQL + Llama3-8B & - & 20.4 & 24.6 & 48.7 & 39.3 & 47.4 \\
DIN-SQL + Qwen2.5-7B & - & 30.1 & 32.4 & 72.1 & 61.2 & 71.1 \\
MAC-SQL +  Llama3-8B & - & 40.7 & 40.8 & 64.3 & 52.8 & 65.2 \\
MAC-SQL + Qwen2.5-7B & - & 46.7 & 49.8 & 71.7 & 61.9 & 72.9 \\

\midrule
\multicolumn{7}{l}{\emph{Fine-Tuning with Open-Source LLMs }}           \\
DTS-SQL-7B~\citep{pourreza2024dts} & 7K & 55.8 & 60.3 & \ 82.7 & \ 78.4 & \  82.8 \\
C{\scriptsize ODE}S-7B~\citep{li2024codes} & - & 57.2 & 58.8 & 85.4 & 80.3 & - \\ 
C{\scriptsize ODE}S-15B~\citep{li2024codes} & - & 58.5 & 56.7 & 84.9 & 79.4 & - \\
S{\scriptsize ENSE}-7B~\citep{yang2024synthesizing} & 25K & 51.8 & 59.3 & 83.2 & 81.7 & 83.5 \\
R{\scriptsize OUTE} + Llama3-8B~\citep{qin2025route}  & 46K & 57.3 & 60.1 & 86.0 & 80.3 & 83.9 \\ 
R{\scriptsize OUTE} + Qwen2.5-7B~\citep{qin2025route} & 46K & 55.9 & 57.4 & 83.6 & 77.5 & 83.7 \\
OmniSQL-7B~\citep{li2025omnisql} & 2.5M & 63.9 & - & - & 81.2 & 87.9  \\
SQLFLOW~\citep{cai2025text2sql} & 90K & 59.2 & - & - & 82.0 & 84.8  \\
\midrule
\rowcolor{gray!10}
\textbf{Ours (\method-Llama-8B)} & 140K & 61.5 & 62.6 & 84.3 & 78.3 & 84.9   \\
\rowcolor{gray!10}
\textbf{Ours (\method-Qwen-7B)} & 140K & 65.1 & 69.6 & 86.1 & 79.7 & 86.1  \\
\bottomrule
\end{tabular}
}
\end{center}
\caption{Main results on BIRD and Spider benchmarks.}
\label{tab_main}
\vspace{-5mm}
\end{table*}

\vspace{-1mm}
\paragraph{Schema-Aware Deduplication.}
Although diversity is encouraged throughout the evolution process, our incremental pipeline can lead to semantic redundancy. Successive transformations may cause different trajectories to converge on similar intents, or mutated queries to be semantically close to their predecessors.
To mitigate this, we perform \emph{schema-aware deduplication}, enforcing diversity independently within each database schema. Specifically, for queries ${q_i}$ associated with the same schema $\mathcal{S}$, we compute semantic representations of their natural language questions using a pretrained encoder and remove samples whose cosine similarity with an existing query exceeds a predefined threshold $\tau$. This process ultimately yields the synthesized dataset $\mathcal{D}_{final}$.

\vspace{-1.5mm}
\paragraph{Supervised Fine-tuning.}

We perform SFT on a base model using $\mathcal{D}_{final}$, training it to generate the reasoning trace $c$ before the target SQL $s$ to internalize structured reasoning patterns.
Specifically, we fine-tune an LLM using a standard cross-entropy objective:
\vspace{-1mm}
\begin{equation}
    \mathcal{L}_{\text{SFT}}(\theta) = -\mathbb{E}_{\mathcal{D}_{final}} \left[ \log \pi_{\theta}(c, s \mid q, \mathcal{S}) \right]
    \vspace{-1mm}
\end{equation}
where $\pi_{\theta}$ represents the initial base model. By exposing the model to execution-verified reasoning trajectories, SFT encourages better generalization to complex and compositional queries.

\section{Experiments}

\subsection{Experimental Setup}
\paragraph{Benchmarks and Metrics.}

We conduct experiments on two primary benchmarks: BIRD~\citep{li2024can} and Spider~\citep{yu2018spider}. 
To further evaluate model robustness and domain generalization, we employ five additional datasets: Spider-DK, Spider-Syn, Spider-Realistic, EHRSQL, and Science Benchmark~\citep{gan2021exploring,gan2021towards,deng2021structure,lee2022ehrsql,zhang2023sciencebenchmark}.
Following prior works, we report Execution Accuracy (EX) as the primary metric. 
For Spider and its variants (Syn, Realistic), we additionally report Test Suite Accuracy (TS)~\citep{zhong2020semantic} to minimize false positives. 
For BIRD, we also include the Valid Efficiency Score (VES)~\citep{li2024can}.
Detailed statistics of all benchmarks and metric definitions are provided in Appendix~\ref{app:datasets}.

\vspace{-1mm}
\paragraph{Baselines.}
We compare \method with three categories of baselines: (i) closed-source prompting methods, (ii) open-source foundation models, and (iii) open-source fine-tuned Text-to-SQL systems. Detailed model lists and configurations are provided in Appendix~\ref{app:baseline}. For fairness, we focus on single-model SFT and exclude reinforcement learning or multi-agent approaches due to their different training and inference complexities.

\begin{table*}[t]
    \centering
    \vspace{-3.5mm}
    \setlength{\tabcolsep}{3pt}
    \renewcommand{\arraystretch}{1.0}
    \resizebox{\textwidth}{!}{
    \begin{tabular}{l|cccccccccc}
      \toprule
      \multirow{2}{*}{Dataset} & \multirow{2}{*}{\# Samples} & \multicolumn{9}{c}{Average Feature Count per SQL} \\
      \cmidrule(l){3-11}
      & & \# Tables. & \# Joins & \# Func. & \# Toks. & \# Agg. & \# Subs. & \# Wins. & \# CTEs & \# Nest. \\
       \midrule
        Spider train~\citep{yu2018spider} & 7000 & 1.69 & 0.54 & 0.65 & 15.88 & 0.53 & 0.15 & 0 & 0 & 1.07 \\
        BIRD train~\citep{li2024can} & 9428 & 2.08 & 1.02 & 1.63 & 25.80 & 0.61 & 0.09 & 0.00 & 0 & 1.08 \\
        \midrule
        \textit{EQE} & 52859 & 2.93 & 1.76 & 2.58 & 33.00 & 0.83 & 0.21 & 0.00 & 0.05 & 1.15 \\
        \textit{OGE-1} & 51646 & 4.04 & 2.49 & 4.60 & 50.74 & 1.23 & 0.62 & 0.01 & 0.25 & 1.38 \\
        \textit{OGE-2} & 24763 & 5.56 & 3.35 & 7.00 & 73.82 & 1.76 & 1.33 & 0.04 & 0.58 & 1.65 \\
        \midrule
        \rowcolor{gray!10}
        \method & 129268 & 3.88 & 2.35 & 4.24 & 47.91 & 1.17 & 0.59 & 0.01 & 0.23 & 1.34 \\
    \bottomrule
    \end{tabular}
    }
    \vspace{-2mm}
    \caption{Comparison of SQL complexity. Metrics indicate the mean frequency of features per SQL. ``Agg.'', ``Func.'', ``Toks.'', ``Subs.'', ``Wins.'', ``CTEs'', and ``Nest.'' denote Aggregates, Functions, Tokens, Subqueries, Window functions, Common Table Expressions, and Nesting levels, respectively. ``OGE-1'' and ``OGE-2'' denote the first and second rounds of Operator-Guided SQL Evolution, respectively.}
    \vspace{-4mm}
    \label{tab:sql_statistics}
\end{table*}

\vspace{-1mm}
\paragraph{Implementation Details.}
Data synthesis utilizes Qwen2.5-Coder-32B-Instruct~\citep{hui2024qwen2} as the evolution and refinement model and Qwen3-Coder-30B-A3B-Instruct~\citep{yang2025qwen3} for reasoning synthesis. Specifically, we execute two rounds of OGE phase. The final training set combines our synthesized dataset with the original BIRD and Spider training sets, all augmented with execution-verified reasoning traces. We then conduct full-parameter SFT on Qwen2.5-Coder-7B-Instruct~\citep{yang2024qwen2} and Meta-Llama-3.1-8B-Instruct~\citep{grattafiori2024llama}, denoted as \method-Qwen-7B and \method-Llama-8B, respectively.  Full implementation details and all prompt templates are available in Appendix~\ref{app:impl} and \ref{app:prompt}, respectively.

\subsection{Main Results}
\label{sec:main_results}

Table~\ref{tab_main} summarizes the performance of \method across BIRD and Spider benchmarks. Our framework demonstrates a significant advantage, particularly on the challenging BIRD dataset. Specifically, \method-Qwen-7B achieves an execution accuracy of 65.1\% on BIRD. This result not only substantially outperforms previous SFT baselines such as SENSE-7B (51.8\%) but also surpasses OmniSQL-7B (63.9\%), which relies on a massive 2.5M synthetic dataset. Notably, \method attains these gains using only approximately 1/18 of the training volume employed by OmniSQL, underscoring the superior information density and quality of our evolutionary data.

On Spider benchmark, \method achieves a competitive 86.1\% EX on the test set, outperforming recent specialized SFT models such as SQLFLOW and SENSE-7B. Notably, our evolutionary synthesis was conducted exclusively using BIRD schemas and seeds. The performance gains on Spider demonstrate that \method effectively instills general structural reasoning capabilities. This confirms that the complexity and logical depth introduced by our evolution are domain-agnostic, providing a robust foundation for real-world generalization even on benchmarks not involved in the synthesis process.

We further assess the impact of \method across different model architectures. For the code-specialized Qwen2.5-Coder-7B, fine-tuning with our data yields a remarkable +13.7\% absolute improvement in EX on BIRD. Similarly, for the general-purpose Llama-3.1-8B-Instruct, our method boosts performance from 42.0\% to 61.5\%, demonstrating strong cross-backbone robustness. Remarkably, our 7B models even surpass sophisticated GPT-4 based pipelines like MCS-SQL (63.4\%), effectively bridging the gap between open-source models and proprietary systems through high-quality data.

% \vspace{-1mm}
\subsection{Analysis of Synthetic Data}
We provide an analysis of our \method dataset evolved from BIRD, characterizing it through semantic diversity and structural complexity (see Appendix~\ref{app:stats} for additional length statistics).

\begin{figure}[t!]
\centering
\includegraphics[width=\columnwidth]{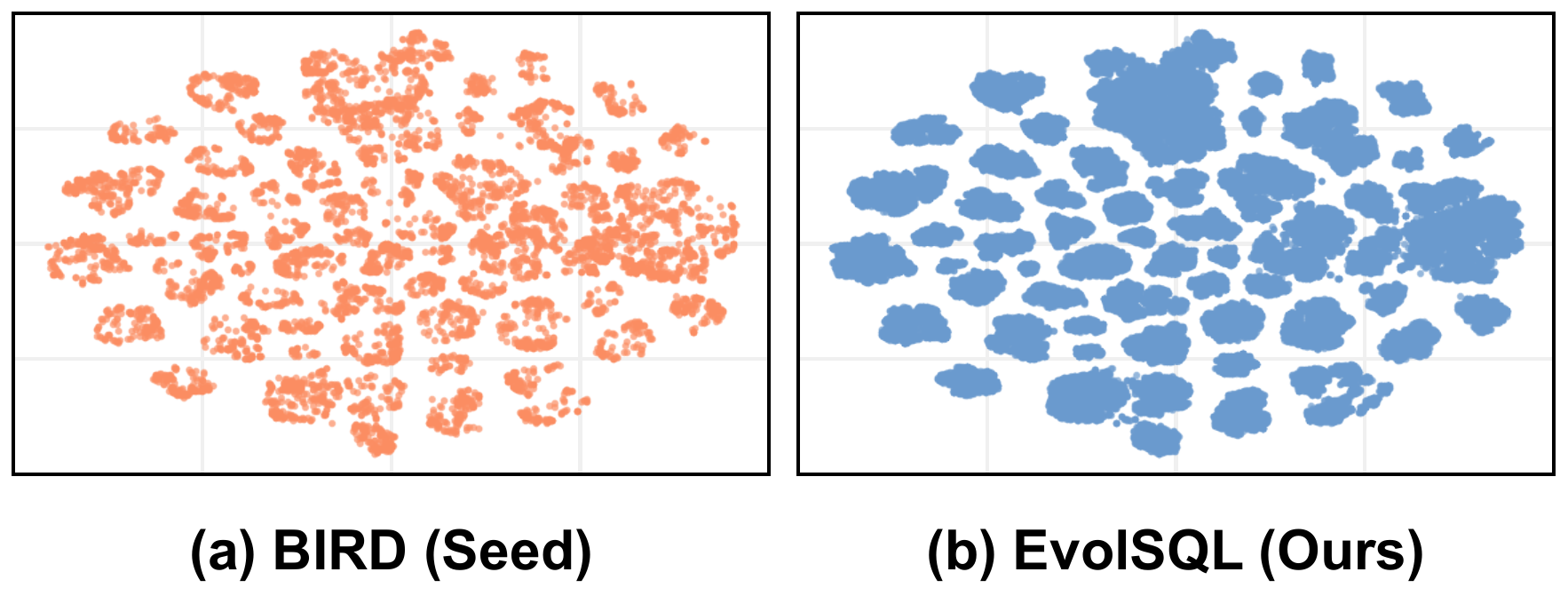}
\vspace{-6mm}
\caption{Comparison of t-SNE visualization between original BIRD train set and \method.}
\vspace{-5mm}
\label{fig:comparison}
\end{figure}

\vspace{-1mm}
\paragraph{Semantic Diversity and Coverage.}
Figure~\ref{fig:comparison} visualizes the distribution of natural user queries in the original BIRD dataset, which exhibits fragmented clusters with noticeable gaps, indicating insufficient coverage over the underlying database schemas. In contrast, \method populates these sparse regions, producing a denser and more continuous semantic landscape. This demonstrates that our framework creates diverse query variants that bridge the semantic gaps present in human-annotated data, thereby offering substantially broader coverage of potential user intents and schema interactions.

\begin{table*}[t!]
  \centering
  \small
  \setlength{\tabcolsep}{6pt}
  \resizebox{\linewidth}{!}{
  \begin{tabular}{lcccccc}
    \toprule
    \textbf{Model} & 
    \textbf{Spider-DK} & 
    \textbf{Spider-Syn} & 
    \textbf{Spider-Realistic} & 
    \textbf{EHRSQL} & 
    \textbf{Science Benchmark} &
    \textbf{Avg.} \\
    \midrule
    Base Model & 67.5 & 63.1 & 66.7 & 24.3 & 45.2 & 53.4 \\
    SFT (BIRD+Spider) & 65.8 & 65.9 & 71.9 & 31.4 & 43.8 & 55.8\\
    OmniSQL-7B & 76.1 & 69.7 & 76.2 & 34.9 & 50.2 & 61.4 \\
    \rowcolor{gray!10}
    {\method-7B (Ours)} & {74.2} & {69.2} & {75.8} & \textbf{38.6} & \textbf{51.8} & \textbf{61.9} \\
    \bottomrule
  \end{tabular}
  }
  \vspace{-2mm}
  \caption{Generalization evaluation results. ``Base Model'' means Qwen2.5-Coder-7B-Instruct.}
  \vspace{-1mm}
  \label{tab:robustness_results}
\end{table*}

\vspace{-1.5mm}
\paragraph{Structural Complexity.}
Table~\ref{tab:sql_statistics} compares SQL complexity across benchmarks and evolution stages. \method exhibits significantly higher structural complexity than standard benchmarks; for instance, the average JOINs increases by 130\% over BIRD, and advanced structures like CTEs are introduced. This complexity is cultivated progressively: while the initial EQE stage produces samples with a structural difficulty closely aligned with the original BIRD dataset, subsequent OGE rounds systematically elevate the structural depth. Notably, the frequency of complex components like CTEs and window functions grows substantially through the iterations (e.g., CTEs from 0.02 to 0.58). This validates that our \emph{atomic transformation operators} effectively steer the generation toward sophisticated logic that remains unattainable for non-progressive or one-shot expansion strategies. To intuitively demonstrate the quality and structural diversity of our synthesized data, we present a detailed case study in Appendix~\ref{app:cases}.

\subsection{Discussions}

\begin{table}[t]
    \centering
    \small
    \vspace{-3mm}
    \resizebox{\linewidth}{!}{
    \begin{tabular}{lccc}
        \toprule
        % 3. 手动拆分表头为两行
        \textbf{Training Data} & \multirow{2}{*}{\textbf{BIRD}} & \textbf{Spider} & \textbf{Spider} \\
        \textbf{Configuration} &  & 
        \textbf{Dev} & \textbf{Test} \\
        \midrule
        \method & \textbf{65.1} & \textbf{79.7} & \textbf{86.1} \\
        \midrule
        \emph{w/o Operators} & 64.5 & 76.7 & 84.4 \\
        \emph{w/o OGE} & 62.7 & 77.9 & 85.7 \\
        \emph{w/o OGE \& EQE}  & 57.4 & 77.7 & 82.8 \\
        \emph{w/o Synthesized CoT} & 63.9 & 78.4 & 86.7 \\
        \emph{w/o Deduplication} & 64.7 & 79.6 & 86.0\\ 
        \bottomrule
    \end{tabular}
    }
    \vspace{-2mm}
    \caption{Ablation study.}
    \label{tab:ablation_stages}
    \vspace{-4mm}
\end{table}

\vspace{-1mm}
\paragraph{Ablation Study.}
Table~\ref{tab:ablation_stages} summarizes the ablation results. The significant performance gap in \textit{w/o OGE \& EQE} confirms that seed data alone is insufficient for complex benchmarks. Crucially, removing atomic transformation operators (\textit{w/o Operators}) consistently degrades performance, demonstrating that fine-grained structural manipulation is key to mastering diverse SQL forms. While \emph{OGE} provides directional evolution, training without synthesized CoT leads to a noticeable drop, especially on BIRD, underscoring the value of explicit reasoning paths for intricate logic. Finally, the decline in \emph{w/o Deduplication} setting validates its role in maintaining dataset quality and structural diversity. Overall, these findings verify that the components of \method effectively synergize to provide the semantic breadth and structural depth necessary for Text-to-SQL modeling.

\begin{figure}[t!]
\centering
\vspace{-4mm}
\includegraphics[width=\columnwidth]{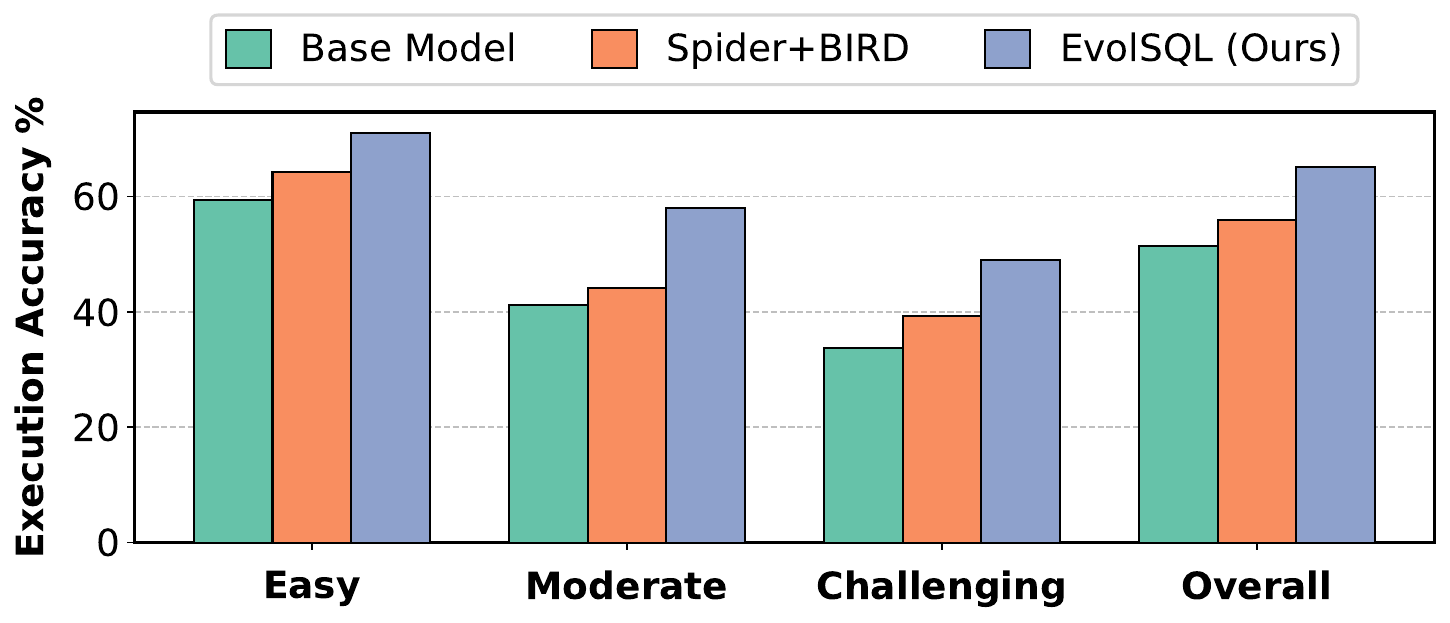}
\vspace{-7mm}
\caption{Execution accuracy (\%) on the BIRD development set across different difficulty levels.}
\vspace{-4.5mm}
\label{fig:bird_difficulty}
\end{figure}

\vspace{-1.5mm}
\paragraph{Performance Analysis across SQL Difficulty.}
Figure~\ref{fig:bird_difficulty} compares BIRD development set performance across difficulty levels. Compared to the baseline (trained on Spider and BIRD), \method achieves consistent improvements, with particularly substantial gains in Moderate (+13.8\%) and Challenging (+9.7\%) subsets. These subsets involve complex joins and logic often under-represented in standard datasets. This indicates that \textit{adaptive directional evolution} effectively synthesizes high-quality complex samples, enabling the model to master intricate SQL logic rather than overfitting to simple patterns.

\vspace{-1.5mm}
\paragraph{Cross-Domain Generalization and Robustness.}
As shown in Table~\ref{tab:robustness_results}, \method consistently outperforms the standard SFT baseline across all tasks, with an average improvement of 6.1\%. Specifically, on Spider-Syn and Spider-Realistic, our model achieves significant gains, indicating its resilience to synonym substitutions and implicit schema mentions scenarios, which are critical for real-world applications. Notably, while OmniSQL-7B shows strong performance on Spider-based variants, \method achieves highly competitive results and even surpasses it on the out-of-domain EHRSQL and Science benchmarks. This suggests that our atomic evolution strategy effectively instills a more domain-agnostic understanding of SQL logic, leading to superior generalization in unseen environments such as healthcare and scientific research. These findings confirm that \method effectively instills a robust understanding of SQL semantics, enabling the model to handle diverse linguistic styles and complex cross-domain requirements.

\vspace{-1mm}
\section{Conclusion}
\vspace{-1mm}
In this paper, we presented \method, a structure-aware data synthesis framework that evolves Text-to-SQL datasets through Atomic Transformation Operators. By decomposing SQL complexity into orthogonal mutations and employing an adaptive directional evolution strategy, \method effectively bridges the gap between simple seed queries and complex real-world applications. On BIRD and Spider, \method matches or exceeds massive-scale synthesis methods while using only 1/18 of the training data volume. Its superior performance on robustness benchmarks further confirms that our evolutionary approach instills  generalized reasoning capabilities that transcend specific database domains.

\section*{Limitations}
Despite the effectiveness of \method, our work has the following limitations:

First, although we incorporate an execution-grounded refinement module to ensure SQL validity, the synthesized dataset may still contain a certain degree of label noise. For instance, a synthesized SQL query might yield the correct execution result by coincidence while its logic slightly deviates from the natural language intent. However, our experimental results suggest that the structural diversity and scale provided by \method effectively outweigh the impact of such minor noise, still leading to high-performance models.

Second, due to resource constraints, we primarily utilized medium-sized open-source models as the evolution and teacher models for data synthesis. While this demonstrates the accessibility of our framework within the open-source ecosystem, it is plausible that employing more powerful proprietary models as teachers could further enhance the quality of reasoning traces and the complexity of the synthesized SQL. We leave the exploration of using stronger teacher models for future work.

Finally, to ensure a fair and direct assessment of the synthesized dataset's quality, our evaluation protocol relies exclusively on Supervised Fine-Tuning. We did not incorporate Reinforcement Learning techniques, which are increasingly common recently in Text-to-SQL task. Nevertheless, we believe that combining our high-quality evolutionary data with RL-based training paradigms could yield further performance gains, representing a promising direction for future research.

\bibliography{custom}

% \bibliography{main}
% \bibliographystyle{acl_natbib}
% \bibliography{main}

% \clearpage
\clearpage
\newpage
\appendix

\appendix
\clearpage

{\raggedbottom

\newtcolorbox{casestudy}[1][]{
  colback=mainboxbg,
  colframe=mainboxborder,
  fonttitle=\bfseries,
  title=Case Study,        
  #1     
}

\section{Algorithm for Operator-Guided SQL Evolution}
\label{app:alg}
\begin{algorithm}[h!]
\caption{Adaptive Directional Evolution}
\label{alg:vertical_evo}
\textbf{Input:} 
    Initial expanded dataset $\mathcal{D}_{H}$, Database schema $\mathcal{S}$, Database $\mathcal{DB}$, Operator family $\Phi$, Strategy model $\mathcal{M}_{strat}$, Rounds $T$, Budget $K$. \\
\textbf{Output:} Final evolved dataset $\mathcal{D}_{evolved}$.

\begin{algorithmic}[1]
\State $\mathcal{D}_{evolved} \leftarrow \mathcal{D}_{H}$, $\mathcal{D}_{curr} \leftarrow \mathcal{D}_{H}$
\State $C(\phi) \leftarrow 0, \forall \phi \in \Phi$; \quad $N_{total} \leftarrow 0$

\For{$t = 1$ to $T$}
    \State $\mathcal{D}_{next} \leftarrow \emptyset$
    \For{each $(q, s) \in \mathcal{D}_{curr}$}
        \State $U_{list} \leftarrow \emptyset$
        \For{each $\phi \in \Phi$}
            \State $S_{feas} \leftarrow \mathcal{M}_{strat}(q, s, \mathcal{S}; \phi)$
            
            \State $P_{accum} \leftarrow C(\phi) / (N_{total} + \epsilon)$ 
            \State $P_{target} \leftarrow 1 / |\Phi|$ 
            \State $W_{div} \leftarrow P_{target} / (P_{accum} + \epsilon)$
            
            \State $U(\phi) \leftarrow S_{feas} \cdot W_{div}$
            \State $U_{list}.\text{add}((\phi, U(\phi)))$
        \EndFor
        
        \State $\Phi^* \leftarrow \text{Top-K}(U_{list}, K)$
        
        \For{each $\phi^* \in \Phi^*$}
            \State $(\tilde{q}, \tilde{s}) \leftarrow \mathcal{M}_{gen}(q, s, \mathcal{S}; \mathcal{I}_{\phi^*})$
            \State $r \leftarrow \text{Exec}(\tilde{s}, \mathcal{DB})$
            \State $s' \leftarrow \mathcal{M}_{refine}(\tilde{q}, \tilde{s}, \mathcal{S}; r)$
            
            \If{$s'$ is valid and result is non-empty}
                \State $\mathcal{D}_{next}.\text{add}((\tilde{q}, s'))$
                \State $\mathcal{D}_{evolved}.\text{add}((\tilde{q}, s'))$
                \State $C(\phi^*) \leftarrow C(\phi^*) + 1$
                \State $N_{total} \leftarrow N_{total} + 1$
            \EndIf
        \EndFor
    \EndFor
    \State $\mathcal{D}_{curr} \leftarrow \mathcal{D}_{next}$
\EndFor
\State \Return $\mathcal{D}_{evolved}$
\end{algorithmic}
\end{algorithm}

\section{Dataset Statistics and Descriptions}
\label{app:datasets}

We evaluate \method on seven benchmarks to comprehensively assess its performance, robustness, and generalization. Our primary targets are Spider\citep{yu2018spider} and BIRD\citep{li2024can}, both designed for cross-domain evaluation where test databases are unseen during training. We report results on the Spider development (1,034 samples) and hidden test sets (2,147 samples), and the BIRD development set (1,534 samples).

To examine resilience to linguistic and knowledge variations, we employ three Spider variants. Spider-DK\citep{gan2021exploring} (535 samples) tests the integration of implicit domain knowledge. Spider-Syn\citep{gan2021towards} (1,034 samples) and Spider-Realistic~\citep{deng2021structure} (508 samples) introduce lexical perturbations, replacing explicit schema mentions with synonyms or implicit references to simulate real-world linguistic variability.

Finally, we assess zero-shot generalization in specialized domains using EHRSQL\citep{lee2022ehrsql} and Science Benchmark\citep{zhang2023sciencebenchmark}. EHRSQL contains 1,008 samples focused on electronic health records (EHR), while Science Benchmark includes 299 samples covering disciplines such as astrophysics and cancer research. As these domains are excluded from our training, they serve as rigorous evaluation for domain-agnostic SQL reasoning.

\section{Baseline Details}
\label{app:baseline}

We provide a comprehensive list of the baselines used in our experiments, categorized into three groups. The \textbf{Proprietary LLM prompting} category includes GPT-4~\citep{achiam2023gpt} evaluated under several prompting frameworks, namely DIN-SQL~\citep{pourreza2023din}, DAIL-SQL~\citep{gao2024text}, MAC-SQL~\citep{wang2025mac}, and MCS-SQL~\citep{lee2024mcs}. 

The \textbf{Open-source models} category covers the zero-shot and few-shot performance of general-purpose foundation models, including Llama3-8B~\citep{touvron2023llama}, Llama-3.1-8B-Instruct~\citep{grattafiori2024llama}, and Qwen2.5-7B~\citep{yang2024qwen2}, as well as the code-specialized Qwen2.5-Coder-7B-Instruct. To further assess their reasoning potential, we also evaluate these models under complex prompting pipelines such as DIN-SQL and MAC-SQL.

The \textbf{Open-source fine-tuning} category comprises specialized Text-to-SQL models and frameworks, including DTS-SQL~\citep{pourreza2024dts}, CODES~\citep{li2024codes}, and ROUTE~\citep{qin2025route}. We also compare against models trained on large-scale synthetic datasets, such as SENSE~\citep{yang2024synthesizing}, OmniSQL~\citep{li2025omnisql}, and SQLFLOW~\citep{cai2025text2sql}. This allows for a direct comparison of data efficiency and performance across different synthesis paradigms.

\section{Implementation Details}
\label{app:impl}

Our data synthesis process is primarily conducted using the schemas and seeds from the BIRD training set, employing Qwen2.5-Coder-32B-Instruct~\citep{hui2024qwen2} as the evolution and refinement model. Specifically, we perform the Operator-Guided SQL Evolution phase for two iterations to progressively enhance structural complexity. To ensure the quality of reasoning traces, we utilize Qwen3-Coder-30B-A3B-Instruct~\citep{yang2025qwen3} as the teacher model to synthesize Chain-of-Thought (CoT) reasoning paths via rejection sampling with $n=4$. During the Schema-Aware Deduplication phase, we apply a semantic similarity threshold of $\tau=0.9$ using the \texttt{all-mpnet-base-v2} encoder. The final training corpus combines our synthesized dataset with the original training sets of BIRD and Spider, both of which are also augmented with execution-verified CoT reasoning. 

For model training, we conduct full-parameter supervised fine-tuning on Qwen2.5-Coder-7B-Instruct~\citep{yang2024qwen2} and Meta-Llama-3.1-8B-Instruct~\citep{grattafiori2024llama} using 8 NVIDIA A100 GPUs. We utilize the AdamW optimizer~\citep{loshchilov2017decoupled} with a peak learning rate of $2 \times 10^{-5}$, a weight decay of 0.1, and a cosine decay schedule with a linear warmup covering the initial 5\% of training steps. We set the global batch size to 512 and train the models for 2 epochs using bfloat16 mixed precision.

\section{Additional Dataset Analysis}
\label{app:stats}

\paragraph{Length Statistics.}

\begin{figure}[h!]
\centering
\includegraphics[width=\columnwidth]{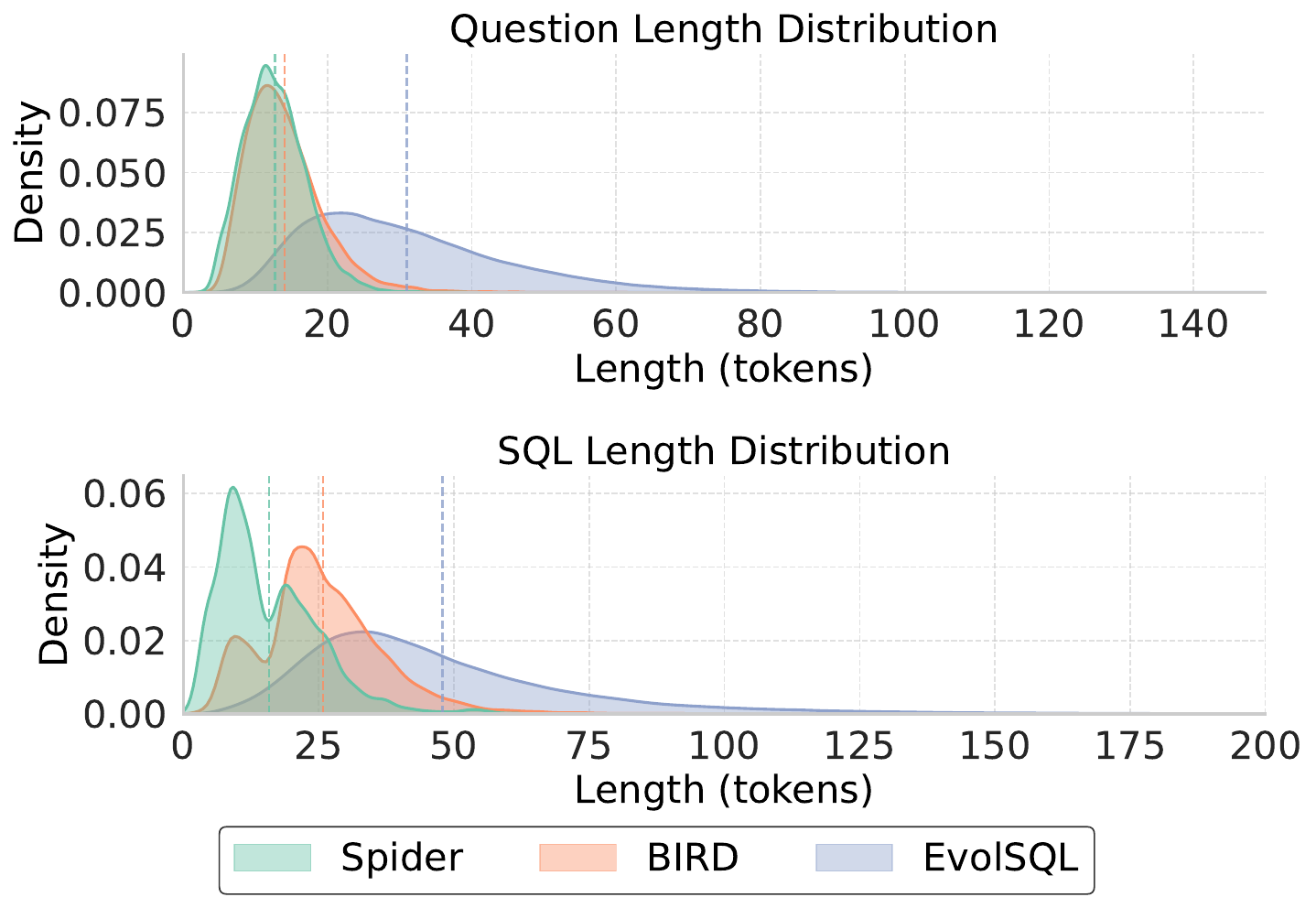}
\caption{Token length distributions for questions and SQL queries in Spider, BIRD, and \method datasets. }
\label{fig:length_distribution}
\end{figure}

Figure~\ref{fig:length_distribution} compares the token length distributions of natural language questions (NL) and SQL queries across Spider, BIRD, and \method.
The NL and SQL distributions for \method exhibit a pronounced shift toward longer sequences compared to BIRD. This increased length serves as a reliable proxy for structural complexity, confirming that our evolutionary framework successfully synthesizes SQL queries with greater depth and intricacy.

\section{Case Study}
\label{app:cases}

\begin{table}[h!]
\small
\centering

\newcommand{\smallsql}[1]{%
    \begingroup\ttfamily\scriptsize\setlength{\baselineskip}{1.5em}#1\endgroup%
}

\newcommand{\newcode}[1]{\textcolor{darkblue}{#1}}

\caption{\textbf{Case Study: An Evolutionary Trajectory.}
An example illustrating how a seed query from BIRD evolves into a more complex SQL structure through adaptive structure-aware evolution.
\textcolor{darkblue}{\textbf{Blue text}} highlights newly introduced SQL components at each evolution stage.}

\resizebox{0.49\textwidth}{!}{
\begin{tabularx}{0.53\textwidth}{X}
\toprule

\textbf{Stage 0: Seed Query (from BIRD)} \\
\textit{Q: Who is the heaviest athlete?} \\
\vspace{1pt}
\smallsql{SELECT full\_name FROM person ORDER BY weight DESC LIMIT 1} \\

\midrule

\textbf{Stage 1: Exploratory Query--SQL Expansion} \\
\textit{Q: Which competitor has won the most medals in the Olympics?} \\
\vspace{1pt}
\smallsql{SELECT p.full\_name FROM person p
JOIN games\_competitor gc ON p.id = gc.person\_id
JOIN competitor\_event ce ON gc.id = ce.competitor\_id
GROUP BY p.id
ORDER BY COUNT(ce.medal\_id) DESC LIMIT 1;} \\

\midrule

\textbf{Stage 2: Operator-Guided SQL Evolution}
\hfill \textit{\textcolor{gray}{(Relational Expansion)}} \\
\textit{Q: Which competitor has won the most medals in the Olympics \textbf{in the sport of Swimming during the Summer season}?} \\
\vspace{1pt}
\smallsql{SELECT p.full\_name FROM person p
JOIN games\_competitor gc ON p.id = gc.person\_id
JOIN competitor\_event ce ON gc.id = ce.competitor\_id
\newcode{JOIN event e ON ce.event\_id = e.id
JOIN sport s ON e.sport\_id = s.id
JOIN games g ON gc.games\_id = g.id
WHERE s.sport\_name = 'Swimming' AND g.season = 'Summer'}
GROUP BY p.id
ORDER BY COUNT(ce.medal\_id) DESC LIMIT 1;} \\

\midrule

\textbf{Stage 3: Operator-Guided SQL Evolution}
\hfill \textit{\textcolor{gray}{(Logical Clause Expansion)}} \\
\textit{Q: Who is the top medal winner in Summer Olympic Swimming?
\textbf{Only consider athletes with at least 3 medals, and use their average winning age to break ties (favoring the youngest).}} \\
\vspace{1pt}
\smallsql{SELECT p.full\_name\newcode{, AVG(gc.age) AS avg\_age}
FROM person p
JOIN games\_competitor gc ON p.id = gc.person\_id
JOIN competitor\_event ce ON gc.id = ce.competitor\_id
JOIN event e ON ce.event\_id = e.id
JOIN sport s ON e.sport\_id = s.id
JOIN games g ON gc.games\_id = g.id
WHERE s.sport\_name = 'Swimming' AND g.season = 'Summer'
GROUP BY p.id
\newcode{HAVING COUNT(ce.medal\_id) >= 3}
ORDER BY COUNT(ce.medal\_id) DESC\newcode{, avg\_age ASC}
LIMIT 1;} \\

\bottomrule
\end{tabularx}
}
\label{tab:case_study}
\end{table}

To provide a concrete understanding of our data synthesis pipeline, we present a representative evolutionary trajectory in Table~\ref{tab:case_study}. The process initiates with a simple seed query from BIRD dataset, which involves a single table and basic sorting logic.

In the Exploratory Query-SQL Expansion phase, the framework diversifies the user intent from querying physical attributes (``heaviest'') to analyzing historical performance (``most medals''). This step effectively broadens the semantic coverage and establishes a multi-table SQL skeleton.
Subsequently, the Operator-Guided SQL Evolution progressively deepens the structural complexity through specific Atomic Transformation Operators.
First, guided by the \emph{Relational Expansion} ($\phi_\texttt{join}$) operator, the query incorporates specific domain constraints (``Swimming'', ``Summer''). As highlighted in \textcolor{darkblue}{\textbf{blue}}, this necessitates the inclusion of three additional tables (`event`, `sport`, `games`) and corresponding join predicates, significantly elevating the relational complexity.
Next, the \emph{Logical Clause Expansion} ($\phi_\texttt{logic}$) operator introduces advanced reasoning requirements. The query is refined to filter aggregated groups (``at least 3 medals'') and apply tie-breaking logic (``youngest on average''). This results in the injection of \texttt{HAVING} clauses and multi-column \texttt{ORDER BY} operations, further enhancing the logical depth of the query.

The final synthesized sample exhibits a high level of complexity by incorporating multi-hop joins, aggregation filtering, and complex sorting, features that are absent in the initial seed. This trajectory validates that our framework offers an effective approach to systematically scale structural complexity and construct high-quality training data.

\clearpage
\onecolumn

\section{Prompts for Text-to-SQL Data Synthesis}
\label{app:prompt}

\subsection{Prompt Template for Exploratory Query-SQL Expansion}
\label{app:eqe}

\setlength{\abovecaptionskip}{0.1cm}
\begin{tcolorbox}[
  colback=yellow!6!white,
  colframe=cyan!50!blue,
  boxrule=0.6pt,
  arc=2pt,
  left=6pt,
  right=6pt,
  top=6pt,
  bottom=6pt
]

Below, you are provided with a database schema and a NL2SQL question. Your task is to draw inspiration from the given NL2SQL question to create a brand new question.

\vspace{0.5em}
\textbf{Database Engine:}\\
SQLite

\vspace{0.5em}
\textbf{Database Schema} \\
% \begin{tcolorbox}[
%   colback=gray!5!white,
%   colframe=gray!50,
%   boxrule=0.4pt,
%   arc=2pt
% ]
% \ttfamily\small
\{DATABASE\_SCHEMA\}

% \end{tcolorbox}
This schema describes the database's structure, including tables, columns, primary keys, foreign keys, and any relevant relationships or constraints. The new question must strictly follow this database schema.

\vspace{0.5em}
\textbf{Evidence} \\
% \begin{tcolorbox}[
%   colback=gray!5!white,
%   colframe=gray!50,
%   boxrule=0.4pt,
%   arc=2pt
% ]
% \ttfamily\small
\{EVIDENCE\}

% \end{tcolorbox}

\textbf{Question} \\
% \begin{tcolorbox}[
%   colback=gray!5!white,
%   colframe=gray!50,
%   boxrule=0.4pt,
%   arc=2pt
% ]
% \ttfamily\small
\{QUESTION\}

% \end{tcolorbox}

\textbf{Gold SQL} \\
% \begin{tcolorbox}[
%   colback=gray!5!white,
%   colframe=gray!50,
%   boxrule=0.4pt,
%   arc=2pt
% ]
% \ttfamily\small
\{GOLD\_SQL\}

% \end{tcolorbox}
\vspace{0.5em}

\textbf{Instructions}
\begin{itemize}
  \item The LENGTH and difficulty level of the new NL2SQL question should be similar to the original one.
  \item You need to ensure that the modified gold SQL still complies with SQLite syntax rules.
  \item You also need to modify the evidence to fit with the new question, keeping it as minimal as possible. Include only the information that is strictly necessary to answer the new question.
  \item You must ensure that the new NL2SQL question can be answered using the given database schema, and you are not allowed to update the schema or introduce new tables or columns.
\end{itemize}

\vspace{0.5em}
\textbf{Output Format}

%In your answer, please respond with a JSON object structured as follows:
\begin{tcolorbox}[
  colback=gray!5!white,
  colframe=gray!50,
  boxrule=0.4pt,
  arc=2pt
]
\ttfamily\small
\{\\
\hspace*{1em}"question": "The new question.",\\
\hspace*{1em}"evidence": "The new evidence. If no evidence needed, it is ok to be an empty string."\\
\hspace*{1em}"gold\_sql": "The corresponding gold sql."\\
\}
\end{tcolorbox}

Take a deep breath and think step by step to increase the difficulty of the question.

\end{tcolorbox}
\captionof{figure}{The prompt template for Exploratory Query-SQL Expansion.}

\subsection{Prompt Template for Operator-Guided SQL Evolution}
\label{app:eqe}

\setlength{\abovecaptionskip}{0.1cm}
\begin{tcolorbox}[
  colback=yellow!6!white,
  colframe=cyan!50!blue,
  boxrule=0.6pt,
  arc=2pt,
  left=6pt,
  right=6pt,
  top=6pt,
  bottom=6pt
]

Below, you are provided with a database schema and a NL2SQL question. Your task is to increase the difficulty of the given NL2SQL question a bit based on the given database schema.

\vspace{0.5em}
\textbf{Database Engine:}\\
SQLite

\vspace{0.5em}
\textbf{Database Schema} \\
% \begin{tcolorbox}[
%   colback=gray!5!white,
%   colframe=gray!50,
%   boxrule=0.4pt,
%   arc=2pt
% ]
% \ttfamily\small
\{DATABASE\_SCHEMA\}
% \end{tcolorbox}

This schema describes the database's structure, including tables, columns, primary keys, foreign keys, and any relevant relationships or constraints. The new question after increasing the difficulty must strictly follow this database schema.

\vspace{0.5em}
\textbf{Evidence} \\
% \begin{tcolorbox}[
%   colback=gray!5!white,
%   colframe=gray!50,
%   boxrule=0.4pt,
%   arc=2pt
% ]
% \ttfamily\small
\{EVIDENCE\}

% \end{tcolorbox}

\textbf{Question} \\
% \begin{tcolorbox}[
%   colback=gray!5!white,
%   colframe=gray!50,
%   boxrule=0.4pt,
%   arc=2pt
% ]
% \ttfamily\small
\{QUESTION\}

% \end{tcolorbox}

\textbf{Gold SQL} \\
% \begin{tcolorbox}[
%   colback=gray!5!white,
%   colframe=gray!50,
%   boxrule=0.4pt,
%   arc=2pt
% ]
% \ttfamily\small
\{GOLD\_SQL\}

% \end{tcolorbox}
\vspace{0.5em}

\textbf{Instructions}
\begin{itemize}
  \item You need to ensure that the modified gold SQL still complies with SQLite syntax rules.
  \item You also need to update the evidence to fit with the new question, but keeping it as minimal as possible. Include only the information that is strictly necessary to answer the new question.
  \item You must ensure that the new NL2SQL question can be answered using the given database schema, and you are not allowed to update the schema or introduce new tables or columns.
  \item To increase the difficulty, you may use the following method, but you are not restricted to it: \{OPERATION\}
\end{itemize}

\vspace{0.5em}
\textbf{Output Format}

%In your answer, please respond with a JSON object structured as follows:
\begin{tcolorbox}[
  colback=gray!5!white,
  colframe=gray!50,
  boxrule=0.4pt,
  arc=2pt
]
\ttfamily\small
\{\\
\hspace*{1em}"question": "The new question after increasing the difficulty.",\\
\hspace*{1em}"evidence": "The new evidence. If no evidence needed, it is ok to be an empty string.",\\
\hspace*{1em}"gold\_sql": "The correspond gold sql after increasing the difficulty."\\
\}
\end{tcolorbox}

Take a deep breath and think step by step to increase the difficulty of the question.

\end{tcolorbox}
\captionof{figure}{The prompt template for Operator-Guided SQL Evolution.}

\newpage

\subsection{Evolution Instructions for Atomic Transformation Operators}
\label{app:ato}
% \begin{center}
\vspace*{0pt} 
\noindent

\setlength{\abovecaptionskip}{0.1cm}
\begin{tcolorbox}[
  colback=yellow!6!white,
  colframe=cyan!50!blue,
  boxrule=0.6pt,
  arc=2pt,
  left=6pt,
  right=6pt,
  top=6pt,
  bottom=6pt
]

\begin{description}[leftmargin=1em, itemsep=0.5em]

    \item[\textbf{1. Functional Wrapping ($\phi_\texttt{func}$):}] 
    Integrate SQL functions to process data within the query. You can use aggregate functions, date functions, mathematical functions, or window functions.

    \item[\textbf{2. Operator Mutation ($\phi_\texttt{op}$):}] 
    Use a wider variety of SQL operators in the query. For example, use \texttt{BETWEEN} for range comparisons, \texttt{IN} or \texttt{NOT IN} to filter against a set of values, or \texttt{LIKE} for pattern matching. You can also introduce conditional logic with a \texttt{CASE WHEN} expression in the \texttt{SELECT} or \texttt{ORDER BY} clause.

    \item[\textbf{3. Logical Clause Expansion ($\phi_\texttt{logic}$):}] 
    Increase the logical complexity within existing SQL clauses. For example, combine multiple conditions in the \texttt{WHERE} clause using \texttt{AND}/\texttt{OR}/\texttt{NOT}; if the original SQL has a \texttt{GROUP BY}, add a \texttt{HAVING} clause to filter the aggregated results; or sort by multiple columns in the \texttt{ORDER BY} clause.

    \item[\textbf{4. Relational Expansion ($\phi_\texttt{join}$):}] 
    Increase the number of tables being joined or change the join type (e.g., switching between \texttt{INNER JOIN} and \texttt{LEFT JOIN}) to introduce new data relationships.

    \item[\textbf{5. Nesting Evolution ($\phi_\texttt{nest}$):}] 
    Make the query structure more complex by introducing nested queries, correlated subqueries, or Common Table Expressions (CTEs).

    \item[\textbf{6. Set Composition ($\phi_\texttt{set}$):}] 
    Use set operators (\texttt{UNION}, \texttt{INTERSECT}, \texttt{EXCEPT}) to combine or compare the result sets of two or more queries. Alternatively, use an \texttt{EXISTS} or \texttt{NOT EXISTS} subquery to check for the existence of records that satisfy specific conditions.

\end{description}

\end{tcolorbox}
\captionof{figure}{Evolution Instructions for Atomic Transformation Operators.}

\subsection{Prompt Template for Strategy Model}
\label{app:strtegy}

\setlength{\abovecaptionskip}{0.1cm}
\begin{tcolorbox}[
  colback=yellow!6!white,
  colframe=cyan!50!blue,
  boxrule=0.6pt,
  arc=2pt,
  left=6pt,
  right=6pt,
  top=6pt,
  bottom=6pt
]

Below, you are provided with a database schema and a Text-to-SQL pair. Your task is to act as an expert data scientist, systematically evaluate the feasibility of applying specific evolution operators to increase the query's complexity, and provide a scored assessment for each.

\vspace{0.5em}
\textbf{Database Engine:} SQLite \\
\textbf{Database Schema:} \{DATABASE\_SCHEMA\} \\
\textbf{Question:} \{QUESTION\} \\ 
\textbf{Gold SQL:} \{GOLD\_SQL\}

\vspace{0.5em}
\textbf{Instructions}
\begin{itemize}
    \item Your primary goal is to analyze the provided SQL query and assess the structural feasibility of applying each of the six evolution operators defined below.
    \item Crucially, you are only evaluating the feasibility. Do NOT generate the new question or new SQL in this step.
\end{itemize}

\vspace{0.5em}
\textbf{Available Evolution Operators}
\begin{description}[leftmargin=1em, itemsep=0.5em]

    \item[\textbf{1. Functional Wrapping:}] 
    Integrate SQL functions to process data (e.g., aggregate functions, date functions, mathematical functions, or window functions).

    \item[\textbf{2. Operator Mutation:}] 
    Use a wider variety of SQL operators (e.g., \texttt{BETWEEN}, \texttt{IN}/\texttt{NOT IN}, \texttt{LIKE}, or \texttt{CASE WHEN}).

    \item[\textbf{3. Logical Clause Expansion:}] 
    Increase logical complexity within clauses (e.g., \texttt{AND}/\texttt{OR}/\texttt{NOT} in \texttt{WHERE}, adding \texttt{HAVING} to filter aggregations, or complex \texttt{ORDER BY}).

    \item[\textbf{4. Relational Expansion:}] 
    Increase the number of tables being joined or change the join type (e.g., \texttt{INNER} vs \texttt{LEFT JOIN}) to introduce new data relationships.

    \item[\textbf{5. Nesting Evolution:}] 
    Introduce nested queries, correlated subqueries, or Common Table Expressions (CTEs).

    \item[\textbf{6. Set Composition:}] 
    Use set operators (\texttt{UNION}, \texttt{INTERSECT}, \texttt{EXCEPT}) or existence checks (\texttt{EXISTS}/\texttt{NOT EXISTS}).

\end{description}

\vspace{0.5em}
\textbf{Output Format}

In your answer, please respond with a JSON object structured as follows:
\begin{tcolorbox}[
  colback=gray!5!white,
  colframe=gray!50,
  boxrule=0.4pt,
  arc=2pt
]
\ttfamily\small
[ \\
\hspace*{1em}\{ \\
\hspace*{2em}"operator": "Functional Wrapping",\\
\hspace*{2em}"score": <float between 0.0 and 1.0>,\\
\hspace*{2em}"justification": "Concise reason regarding schema availability."\\
\hspace*{1em}\},\\
\hspace*{1em}...\\
]
\end{tcolorbox}

Take a deep breath and think step by step to evaluate the feasibility of all six operators based on the schema and current SQL.

\end{tcolorbox}
\captionof{figure}{The prompt template for Strategy Model.}

\clearpage
\twocolumn

}

\end{document}